\documentclass{article}

\usepackage[preprint]{neurips_2021}

\usepackage[utf8]{inputenc} 
\usepackage[T1]{fontenc}    
\usepackage{hyperref}       
\usepackage{url}            
\usepackage{booktabs}       
\usepackage{amsfonts}       
\usepackage{nicefrac}       
\usepackage{microtype}      
\usepackage{xcolor}         
\usepackage[pdftex]{graphicx}
\usepackage{natbib}
\setcitestyle{numbers}
\usepackage{amsmath}
\usepackage{amssymb}
\usepackage{enumitem}

\title{Delving into the pixels of adversarial samples}

\author{%
  Blerta Lindqvist \\
  Department of Computer Science\\
  Aalto University\\
  Helsinki, Finland \\
  \texttt{blerta.lindqvist@aalto.fi} \\
}

\begin{document}

\maketitle

\begin{abstract}

Despite extensive research into adversarial attacks, we do not know how adversarial attacks affect image pixels.
Knowing how image pixels are affected by adversarial attacks has the potential to lead us to better adversarial defenses. Motivated by instances that we find where strong attacks do not transfer, we delve into adversarial examples at pixel level to scrutinize how adversarial attacks affect image pixel values. We consider several ImageNet architectures, InceptionV3, VGG19 and ResNet50, as well as several strong attacks. We find that attacks can have different effects at pixel level depending on classifier architecture. In particular, input pre-processing plays a previously overlooked role in the effect that attacks have on pixels. Based on the insights of pixel-level examination, we find new ways to detect some of the strongest current attacks.

\end{abstract}

\section{Introduction}

In this work, we find instances where adversarial attack transferability of strong attacks is low, despite transferability being considered to be pervasive~\cite{szegedy2013intriguing,kurakin2016adversarial,goodfellow6572explaining, carlini2017towards}. In addition, we find that high attack transferability is not always indicative of a strong attack because some attacks can generate out-of-domain images that also misclassify across classifiers. Motivated by these findings, we focus on how attacks change pixels. The examination of pixel changes under attacks leads us to simple ways of detecting some of the strongest current attacks.

Transferability of adversarial attacks is currently measured with transferability metrics~\cite{kurakin2016adversarial,zhou2018transferable,inkawhich2019feature,huang2019enhancing, inkawhich2020transferable,NEURIPS2020_eefc7bfe} that may not convey attack strength as intended. Attack transfer metrics are used to show the rate at which adversarial examples generated in one classifier cause misclassification in another classifier. However, there is an underlying assumption that the generated adversarial images do indeed belong to the dataset domain of images. This may not be the case because many attacks do not even minimize perturbation, using instead parameters that control perturbation: $\kappa$ in Carlini\&Wagner (CW) attack~\cite{carlini2017towards} and $\epsilon$ in Projected Gradient Descent (PGD)~\cite{madry2017towards} and Fast Gradient Sign Method (FGSM)~\cite{goodfellow6572explaining}. In addition, these parameters can cause different amounts of perturbation in different classifiers, depending on classifier architecture. We show that adversarial attacks can result in images that no longer belong in the dataset domain of images. Yet the attack transfer rate for such attacks is still high because other classifiers would also misclassify out-of-domain images. This indicates that we need to look deeper and beyond transferability metrics, into perturbation values, images pixel values and how classifier architecture affects perturbation.

\begin{figure}[ht]
  \centering
  \includegraphics[width=0.75\linewidth]{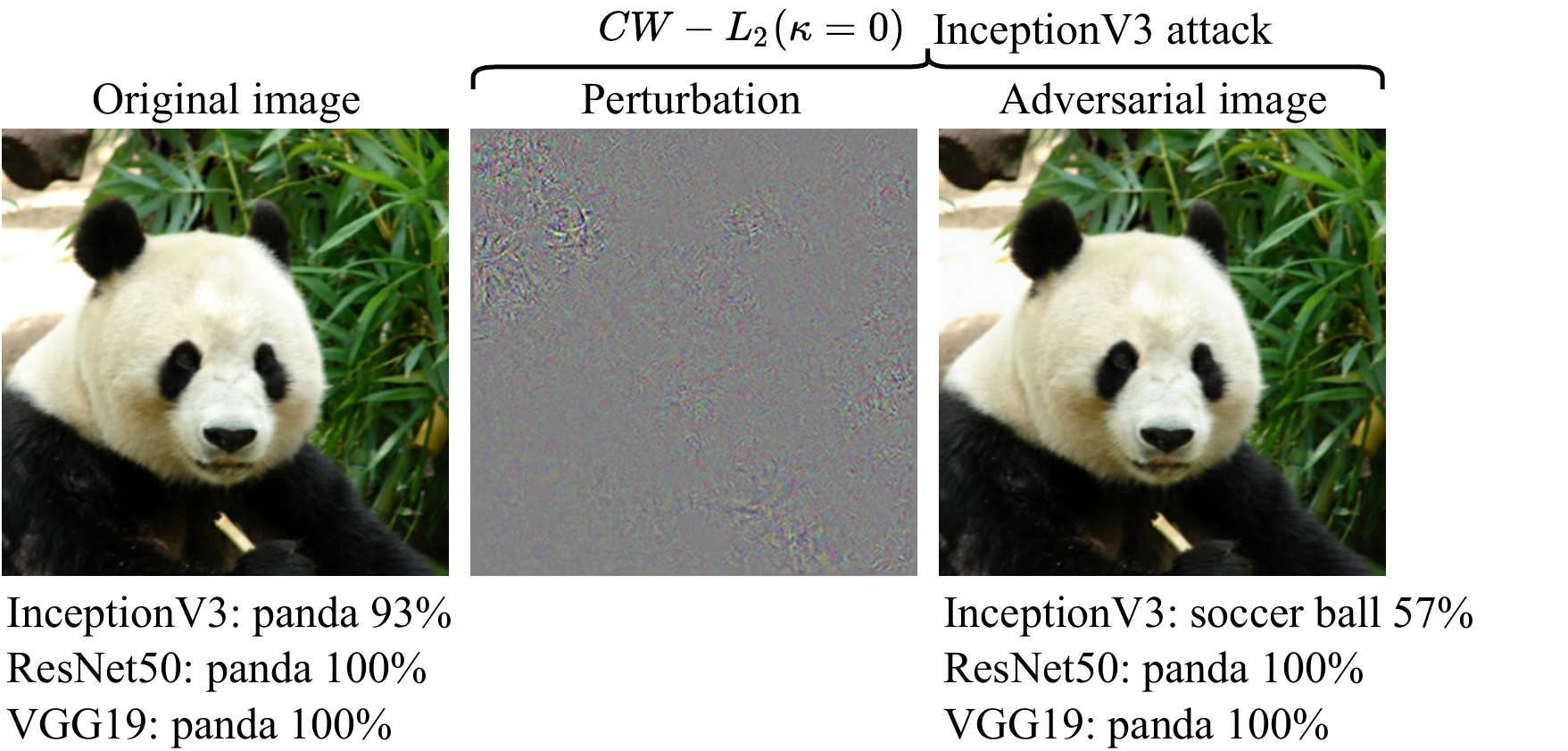}
  \caption{An adversarial attack on InceptionV3 that does not transfer to ResNet50 or VGG19.}
  \label{fig:panda}
\end{figure}

Based on our findings of low attack transferability between classifiers of well-known architectures, a question arises:

\textit{Why has this low attack transferability not been found out before?}

On the side of adversarial attacks, the initial findings of adversarial transferability by~\citet{szegedy2013intriguing,goodfellow6572explaining} examined transferability between simple convolutional and fully-connected models. Despite the advent of new architectures such as such as VGG~\cite{simonyan2015very}, ResNet~\cite{he2016deep} and Inception~\cite{szegedy2017inception}, some adversarial attacks have conducted limited transferability experiments. For example, in CW attack, \citet{carlini2017towards} test transferability to similar classifier, DeepFool~\cite{moosavi2016deepfool} does no transferability tests, \citet{madry2017towards} in PGD attack tests the same architecture with layers of different sizes and different initializations. More recent research~\cite{zhou2018transferable,inkawhich2019feature,huang2019enhancing, inkawhich2020transferable,NEURIPS2020_eefc7bfe} on adversarial attacks is exclusively on the transferability of $L_{\infty}$ attacks between different architectures, but not other norms.

On the side of adversarial defenses, transferability to different architectures may have not been researched for different reasons. The main reason is the establishement of a framework for the evaluation of adversarial defenses~\cite{carlini2019evaluating} that required that defenses should perform transferability experiments with another model that is as similar to the defended model as possible. This framework was initiated by~\citet{carlini2017towards}, based on how \citet{carlini2017towards} defeated Defensive Distillation~\cite{papernot2016distillation}. But even before that framework, \citet{kurakin2016adversarial} test the transferability in the adversarial training defense with variations of the Inception architecture on ImageNet. Having to check transferability to very similar architectures~\cite{carlini2019evaluating}, defenses do not test against transferability to different architectures because of the computational demands of generating adversarial examples.

The main contributions of our work are:
\begin{itemize}
\item Though adversarial attacks are claimed to be highly transferable, we find that even some of the strongest current attacks can have low transfer rates. We illustrate this in Figure~\ref{fig:panda} with a CW-$L_2, \kappa=0$ attack that does not transfer. Further results are in Table~\ref{tbl-imagenet-untargeted}, where transfer rates of CW-$L_2$ and several PGD attacks are as low as $6$\%. Based on this, defenses can use different classifiers to tell apart adversarial samples when the classifiers disagree.

\item We find that transfer rates alone can be misleading indicators of attack strength. Some attacks can have high transfer rates because they generate images that do not belong in the domain of images and thus cause misclassification, such as blank or noise images. Attacks with high transfer rates shown in Table~\ref{tbl-imagenet-untargeted} are CW-$L_{\infty}$ and PGD $\epsilon=8$ against InceptionV3 classifier. Figure~\ref{fig:incv3_pgd} shows that the PGD $\epsilon=8$ attack against InceptionV3 generates noise images. Figure~\ref{fig:cwinf_vgg_resnet_incv3} shows that the highly transferable CW-$L_{\infty}$ attacks are in fact mostly blank.

\item Though we do not set out to do adversarial detection, the insights we gain from looking at what happens to image pixels under attacks enables us to devise simple ways to tell apart original samples and adversarial examples for CW, PGD and FGSM attacks. We show this for PGD and FGSM in Subsection~\ref{sec-resnet-vgg-pgd-fgsm} and Subsection~\ref{sec-incv3-pgd-fgsm}, and for CW-$L_{\infty}$ in Subsection~\ref{sec-resnet-vgg-cwinf} and Subsection~\ref{sec-incv3-cwinf}.

\item We find that image pre-processing can affect perturbation magnitude, shown in Table~\ref{tbl-distances}.

\end{itemize}

\section{Background and related work}

\subsection{Adversarial attacks}

\textbf{Fast Gradient Sign Method} The FGSM attack by~\citet{goodfellow6572explaining}, is a simple way to generate adversarial examples: $x_{adv} = x + \epsilon \cdot sign(\triangledown_x   J(x,y_{true})$, where $J$ is the loss function.

\textbf{Carlini\&Wagner attack} CW~\cite{carlini2017towards} is one of the current strongest white-box attacks. CW formulate their attack as an unconstrained optimization based on the following minimization problem:

\begin{align*}
& {\text{minimize}}
& &  \| x_{adv}-x\| + c \cdot loss_{f}(x_{adv},t)\\
& \text{subject to}
& &  x_{adv} \in [0,1]^n,
\end{align*}

where $\|v\|$ denotes a norm of vector $v$, and where $c$ is a regularizing parameter, t is the target class. The CW attack can work on different perturbation norms, usually $L_2$ and $L_{\infty}$. However, CW does not only minimize perturbation. Instead, it uses a $\kappa$ parameter that controls the confidence of adversarial samples, taking values between $0$ and up to $100$.

\textbf{Projected Gradient Descent attack} The Projected Gradient Descent (PGD) attack~\cite{madry2017towards} is a multi-step attack, where the perturbation of each step is determined by an $\alpha$ parameter. The PGD iterations start at a random point $x_0$ which is then perturbed. The perturbation gets projected on an $L_p$-ball $B$ at each iteration: ${x(j+1)=Proj_B(x(j)+\alpha \cdot sign(\nabla_x loss(\theta,x(j),y))}$.

\subsection{Adversarial defenses}

Adversarial Training (AT)~\cite{szegedy2013intriguing,kurakin2016adversarial,madry2017towards} is the first adversarial defense and the only undefeated defense so far. AT enhances the training data with adversarial examples labelled correctly. One type of defense that has been consistently shown not to work is adversarial detection, with many detection methods defeated~\cite{sabour2015adversarial, carlini2017adversarial, carlini2017magnet, tramer2020adaptive}. Other defenses based on gradient masking and obfuscation have also been defeated~\cite{carlini2016defensive,athalye2018obfuscated}.

\paragraph{ImageNet architectures}
\label{sec-arches}

There are several ImageNet architectures available: VGG~\cite{simonyan2015very}, ResNET~\cite{he2016deep}, Inception~\cite{szegedy2016rethinking}, Xception~\cite{chollet2017xception}, InceptionV3~\cite{szegedy2017inception}, EfficientNet~\cite{tan2019efficientnet}. Architectures can pre-process images differently. In both ResNet~\cite{he2016deep} and VGG~\cite{simonyan2015very}, images are mean-centered, and converted from RGB to BGR format. The subtracted channel mean values are calculated from the ImageNet dataset: $103.939$, $116.779$, $123.68$. In InceptionV3~\cite{szegedy2017inception}, images are normalized to the $0.0-2.0$ range.


\paragraph{Recent transferability research}

There is an unchallenged consensus that transferability is a widespread phenomenon and adversarial samples generated with one classifier are highly likely to transfer to other classifiers~\cite{goodfellow6572explaining,carlini2017towards,chen2017zoo}.
Current transferability research has since focused on increasing the transferability of $L_{\infty}$-norm multi-step attacks in a black-box setting. There have been several $L_{\infty}$ multi-step attacks. For example, Transferable Adversarial Perturbations (TAP)~\cite{zhou2018transferable} - an attack that maximizes distance between natural images and their adversarial samples using intermediate feature maps. Activation Attack (AA)~\cite{inkawhich2019feature} and Intermediate Level attack (ILA)~\cite{huang2019enhancing} are two other attacks that target a single hidden layer but different distortion. \citet{NEURIPS2020_eefc7bfe} extends the FDA~\cite{inkawhich2020transferable} attack do to do multi-layer perturbations. Linear Backpropagation method focuses on untargeted $L_{\infty}$ attacks~\cite{NEURIPS2020_00e26af6}. Overall, recent transferability research has focused only $L_{\infty}$ attacks. Crucially, they focus only on transferability metrics without making sure that the resulting adversarial samples belong to the dataset image domain.

\paragraph{Explanations for adversarial transferability}

There are several hypotheses that attempt to explain adversarial transferability. \citet{ilyas2019adversarial} explain both adversarial attacks and adversarial transferability with non-robust features, claiming that transferability is caused by different models learning similar non-robust features. \citet{madry2017towards} also relate transferability to classifier robustness, and observe transferability decrease with higher classifier capacity. \citet{goodfellow6572explaining} explain transferability with models learning similar model weights, and that perturbations are highly-aligned with the weights. \citet{tramer2017space}, find that it is decision boundary similarity that enables transferability adversarial subspaces. They also find adversarial subspaces and conclude that the higher the dimensionality of these subspaces, the likelier it is that these subspaces will intersect, resulting in transferability. These subspaces resemble the so-called pockets in the manifold with low probability that~\citet{szegedy2013intriguing} explain adversarial attacks with. This view is also supported by \citet{goodfellow6572explaining}, explaining them as a property of high dimensional dot products, and a result of models being too linear, further supported by~\citet{NEURIPS2020_00e26af6}. Even though explanations for adversarial attacks or transferability may differ, one thing they all agree on is that adversarial attacks are highly transferable~\cite{szegedy2013intriguing, goodfellow6572explaining,chen2017zoo, ilyas2019adversarial}.

\section{Methodology}

We experiment on adversarial evasion attacks. We consider cross-model transferability between InceptionV3~\cite{szegedy2017inception}, ResNet50~\cite{he2016deep} and VGG19~\cite{simonyan2015very}. For transferability experiments, we use one classifier to generate adversarial samples and the other classifiers used as target classifiers. We construct adversarial samples from 1000 randomly sampled clean images from the validation set of ImageNet~\cite{ILSVRC15}, as suggested by ~\citet{ILSVRC15}. We conducted indepedent experiments for different adversarial attacks. Transferability was measured using transfer rate as defined by~\citet{kurakin2016adversarial}.

\textbf{Transfer rate} From 1000 randomly sampled clean images, we pick only the misclassified adversarial samples for the source model that had classified correctly as clean images (correctly classified as clean samples but misclassified as adversarial samples by the source model). We then measure the fraction of them that was misclassified by the target classifiers - transfer rate. This transfer rate reflects an attack where the attacker chooses samples that have misclassified in the source classifier and deploys them on the target classifiers.

\textbf{Datasets} We use the ImageNet~\cite{ILSVRC15} in our experiments. ImageNet is a $1000$-class dataset with $1281167$ training images. Evaluation of ImageNet with the validation set has been suggested by ~\citet{ILSVRC15}. For the architectures we use, the size of the ImageNet images used is $224 \times 224 \times 3$ for ResNet50 and VGG19 classifiers, and $299 \times 299 \times 3$ for InceptionV3 classifier.

\textbf{ImageNet pre-trained classifiers} We use pre-trained ImageNet~\cite{ILSVRC15} classifiers with no adversarial defense, with weights loaded from Keras~\cite{chollet2015keras}.

\textbf{Perturbation norm} We use $L_{\infty}$ norm to make image perturbations comparable since the architectures we consider take inputs of different sizes. $L_{\infty}$ norm facilitates comparing the perturbations.

\textbf{Attacks} We generate adversarial examples with white-box attacks on one classifier and measure the attack transfer rate to other classifiers. None of the classifiers employ any adversarial defense. We consider targeted and untargeted evasion attacks and assume that the image perturbations are $L_p$-constrained. We assume that the adversary has access to the ImageNet dataset and is able to manipulate image pixels. We use Adversarial Robustness Toolbox (ART)~\cite{art2018} for FGSM, CW-$L_{inf}$ and PGD attacks, and CleverHans~\ref{papernot2018cleverhans} for CW-$L_{2}$ attack. We use the CW, PGD and FGSM attacks. The settings for CW are: 9 steps, $1K$ iterations, $0$ to $100$ confidence values for CW-$L_2$ and $0$ for CW-$L_{inf}$, based on~\cite{carlini2017towards}. For the PGD attack, the parameters are based on the PGD paper~\cite{madry2017towards}:$7$ steps of size $2$ with a total $\epsilon=8$ for untargeted and $\epsilon=16$ for targeted attacks (based also on ~\cite{kurakin2018adversarial}). Because such $\epsilon$ values result in out-of-domain, just-noise adversarial images, we conduct additional experiments with $\epsilon$ values adjusted to the pixel range of InceptionV3 pre-processing. ResNet50 and VGG19 have a pixel ange of $0$ to $255$, whereas InceptionV3 has a range of $-1$ to $1$. Therefore, we adjust $\epsilon$ values for additional InceptionV3 experiments to $\epsilon=8/127.5$ for untargeted and $\epsilon=16/127.5$ for targeted attacks. For all PGD attacks, we use $0$ random initialisations. For FGSM attacks we use $\epsilon$ values up to $160$ in order to study perturbation convergence. The target labels for targeted attacks are selected uniformly at random.

\section{Transferability and perturbation experiments}
\label{sec-transf-exp}

Here, we conduct experiments by generating adversarial samples using one pre-trained classifier, and evaluating attack transfer rate to other classifiers. Table~\ref{tbl-imagenet-untargeted} summarizes results for untargeted attacks, Table~\ref{tbl-imagenet-targeted} in the Appendix summarizes results for targeted attacks.

\begin{table*}[ht]
  \caption{Here we show transfer rates and average perturbation for untargeted attacks on ImageNet classifiers. (-) indicates that transfer rate could not be measured based on the definition of transfer rate because there was no original sample that classified correctly, the adversarial example of which became adversarial. We make three important observations from the results. (1) The attacks with highest transfer rate are the ones that cause close to maximum adversarial perturbation that classifier pixel range allows ($2.0$ for InceptionV3 and $255.0$ for ResNet50 and VGG19). However, the high transfer rates fail to capture the fact that many of the adversarial examples that transfer to other classifiers are out-of-domain, just-noise images that naturally misclassify. The high perturbation values are an indication that the generated adversarial examples do not look like domain images. (2) The transfer rates of strong attacks such as CW and PGD can be as low as $6$\%. (3) For ResNet50 and VGG19, adversarial perturbation seems to saturate to values well below the maximum that the $255.0$ pixel range should allow. The value where adversarial perturbation seems to saturate to is an indication that adversarial perturbation may depend on input pre-processing. ResNet50 and VGG19 share the same input pre-processing. Targeted attack experiments are shown in Table~\ref{tbl-imagenet-targeted} in the Appendix with similar results.}
  \label{tbl-imagenet-untargeted}
  \centering
  \begin{tabular}{llrrrr}
    \toprule
          \multicolumn{2}{c}{Attack}  & \multicolumn{1}{c}{Perturbation} & \multicolumn{3}{c}{Transfer rate} \\
                                        \cmidrule(r){1-2}   \cmidrule(r){3-3} \cmidrule(r){4-6}
    Attacked    &                   & Avg. $L_{\infty}$   & & & \\
    model       & Attack            & distances & InceptionV3 & VGG19  & ResNet50 \\
    
    \midrule

    InceptionV3 & PGD $\epsilon=8$      & 1.99 & 100\% & 100\% & 100\% \\ 
    InceptionV3 & PGD $L_{\infty}, \epsilon=8/127.5$  & 0.06 & 100\% &  19\% &  14\% \\ 
    InceptionV3 & CW-$L_2, \kappa=0$                & 0.02 & 100\% &  17\% &  12\% \\ 
    InceptionV3 & CW-$L_2, \kappa=100$              & 0.27 & 100\% &  22\% &  17\% \\ 
    InceptionV3 & CW-$L_{\infty}$                   & 1.00 & 100\% &  97\% & 100\% \\ 

    \midrule
    
    VGG19 & PGD $\epsilon=8$         &   8.00 &  34\%          & 100\% & 31\%  \\ 
    VGG19 & CW-$L_2, \kappa=0$      & 2.22 &  6\% & 100\% & 11\% \\ 
    VGG19 & CW-$L_2, \kappa=100$    & - &  - & - & - \\ 
    VGG19 & CW-$L_{\infty}$         & 144.52 & 100\%          & 100\% & 100\%         \\ 

    \midrule
    
    ResNet50 & PGD $\epsilon=8$      &   8.00 & 36\% & 42\% & 100\% \\ 
    ResNet50 & CW-$L_2, \kappa=0$   & 1.68 & 9\% & 17\% & 100\% \\ 
    ResNet50 & CW-$L_2, \kappa=100$ & - & - & - & - \\ 
    ResNet50 & CW-$L_{\infty}$      & 144.35 & 100\% & 100\% & 100\% \\ 

    \bottomrule
  \end{tabular}
\end{table*}

\textbf{Important observations from transferability and perturbation experiments} We make the following observations based on Table~\ref{tbl-imagenet-untargeted} and Table~\ref{tbl-imagenet-targeted}:

\begin{itemize}
\item It is the attacks with perturbation closest to maximum adversarial perturbation ($2.0$ for InceptionV3 and $255.0$ for ResNet50 and VGG19) that have the highest transfer rates. However, the high perturbation also indicates that the generated adversarial examples might be out-of-domain images. We explore and show this in the following sections.
\item We observe that transfer rates of some of the current strongest attacks are as low as $6$\%.
\item Adversarial perturbation in attacks against ResNet50 and VGG19 seems to saturate to similar values that are far from the $255.0$ that should be achieveable based on ResNet50 and VGG19 pixel ranges. We suspect that this might be related to the common input-preprocessing in ResNet50 and VGG19. We explore the limit of perturbation in classifiers in Table~\ref{tbl-distances}.
\end{itemize}

To resolve what the saturation value for attack perturbation is, we conduct FGSM experiments with a range of values. Table~\ref{tbl-distances} shows that the maximum value of perturbation in ResNet50 and VGG19 is directly related to input image pre-processing: $151.061=255.0-103.939$, where $103.939$ is the minimum of channel means that gets subtracted from input images. This shows that input pre-processing directly impacts perturbation and hints at the possibility that there might be some regularity in how attacks change pixels that we might be able to exploit. The following sections look closer at how image pixel values change under different attacks.

\begin{table*}[ht]
  \caption{Here, we show that $L_{\infty}$ perturbation saturates to a value much smaller than $255.0$ in ResNet50 and VGG19. Untargeted FGSM attacks used. We observe that in both ResNet50 and VGG19, the value of $L_{\infty}$ saturation is $151.061$. This value is related to the input pre-processing value of $103.939$ because $151.061=255.0-103.939$. }
  \label{tbl-distances}
  \centering
  \begin{tabular}{rrrrrr}
    \toprule
    \multicolumn{2}{c}{InceptionV3}  & \multicolumn{2}{c}{ResNet50} & \multicolumn{2}{c}{VGG19} \\
            \cmidrule(r){1-2} \cmidrule(r){3-4} \cmidrule(r){5-6}
    FGSM $\epsilon$ & Max $L_{\infty}$ & FGSM $\epsilon$ & Max $L_{\infty}$ & FGSM $\epsilon$ & Max $L_{\infty}$ \\
                    & perturbation &                 & perturbation &                 & perturbation \\
    \midrule
    2.0 & 2.0 & 150.0 & 150.0   & 150.0 & 150.0   \\
    4.0 & 2.0 & 160.0 & \textbf{151.061} & 160.0 & \textbf{151.061} \\
    \bottomrule
  \end{tabular}
\end{table*}

\section{Examination of ResNet50 and VGG19 adversarial examples at pixel level}
\label{sec-resnet-vgg-pixel}

We examine ResNet50 and VGG19 adversarial examples together because pixel changes are similar in adversarial examples generated from these two classifiers. We examine the images in their ResNet50 and VGG19 pre-processed format, as described in Section~\ref{sec-arches}.

\subsection{PGD and FGSM attacks}
\label{sec-resnet-vgg-pgd-fgsm}

\begin{figure}
  \centering
  \includegraphics[width=0.65\linewidth]{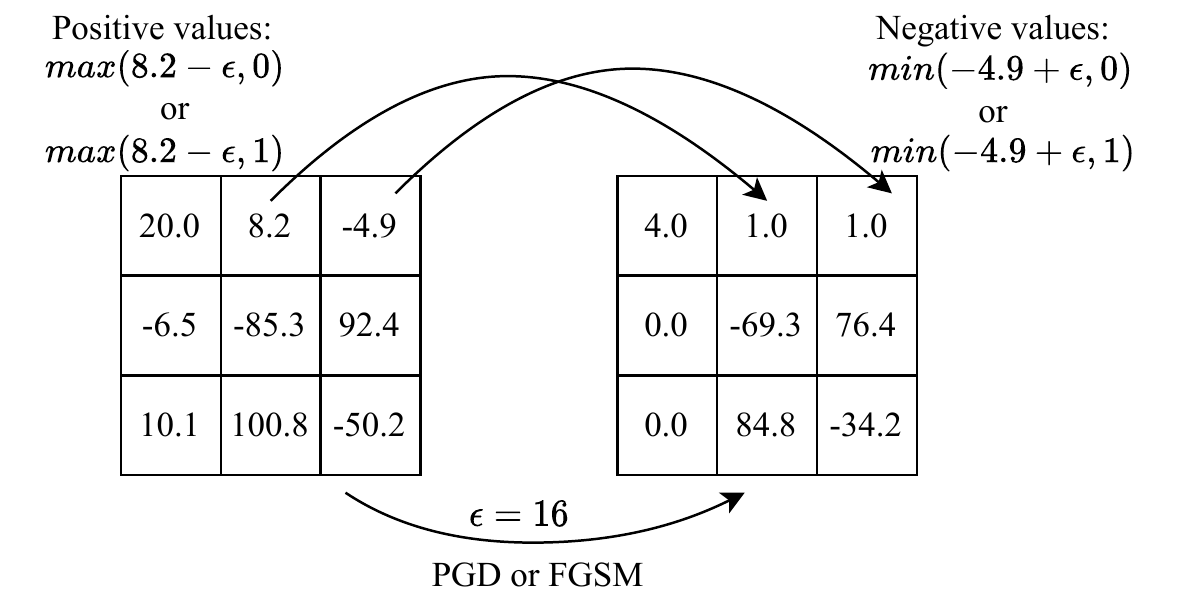}
  \caption{Here, we show how pixel values change when an original image becomes adversarial under PGD and FGSM attack using a $3 \times 3$ image for illustration. Positive pixels decrease in value by $\epsilon$ towards $0$ or $1$, while negative pixels increase in value by $\epsilon$ towards $0$ or $1$.}
  \label{fig:pixel-changes}
\end{figure}

For both PGD and FGSM, pixel values change by $\epsilon$ converging towards $0.0$ or $1.0$. Positive pixel values decrease by $\epsilon$ towards either $0$ or $1$, whereas negative pixel values increase by $\epsilon$ to either $0$ or $1$. Figure~\ref{fig:pixel-changes} illustrates the pixel changes.

As Figure~\ref{fig:pixel-changes} shows, PGD and FGSM attacks move pixel values away from extremum values towards $0$ or $1$ by $\epsilon$. This makes it possible to detect adversarial images based on channel minimum pixel values. Towards that goal, we take a closer look at channel minimum and maximum pixel values in $1000$ random original images and find the following channel-wise pixel information, summarized in Table~\ref{tbl-pgd-min-max} in the Appendix.

In Figure~\ref{fig:channel_min_max}, we summarize channel minimum and maximum values visually, as well as the values where channel pixel minimums and maximums move to after PGD or FGSM attack of $\epsilon=8$ and $\epsilon=16$. The histograms in Figure~\ref{fig:channel_min_max} show that channel extremum pixel values for most of $1000$ random, original images are close to channel limit values, which are calculated as the limit pixel values $0.0$ and $255.0$ after ResNet50 and VGG19 pre-processing. We use this to find original examples and adversarial examples.

\begin{figure}
  \centering
  \includegraphics[width=0.8\linewidth]{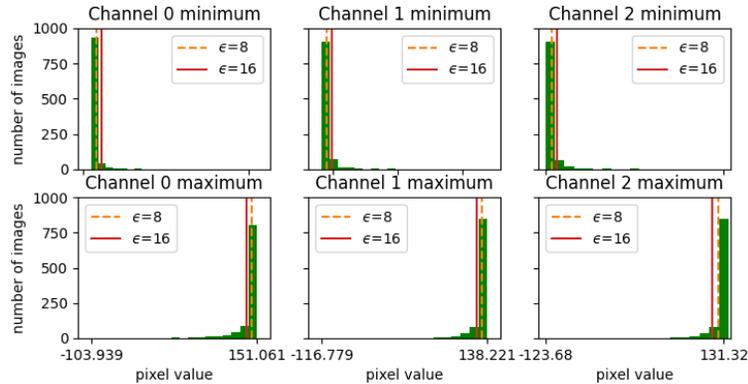}
  \caption{Here, we examine $1000$ original, non-adversarial images to show that for all channels, channel minimum and maximum values are close to channel minimum and channel maximum limits respectively for most images. We can use this to detect when pixel values have been changed with PGD and FGSM attacks by $\epsilon$ towards $0$ or $1$.}
  \label{fig:channel_min_max}
\end{figure}

\textbf{Detection of PGD and FGSM adversarial examples} Based on how PGD and FGSM change image pixels, we can devise a simple way to detect adversarial examples based on $\epsilon$. For example, to detect PGD or FGSM adversarial examples generated with an $\epsilon=8$, we find images where all pixel values have at least an absolute difference of $8$ from channel limit values. This would detect all true positive adversarial examples. but could lead to false positives. We conduct an experiment to find out how many original, non-adversarial images would be falsely detected as adversarial, false positives, based on this detection method. We find that for $\epsilon=8$, $21$ images are falsely detected as adversarial, out of $1000$ original, non-adversarial images. For $\epsilon=16$, only $3$ images are falsely detected as adversarial, out of $1000$ original images.

\subsection{CW-$L_2$ attack} We do not notice any regularity in pixel changes that can lead to detection. However, the non-transferability of the CW attacks to other classifiers makes feasible the following adversarial detection.

\textbf{Detection of CW-$L_2$ adversarial examples} Having two classifiers with different architectures, for example VGG19 and InceptionV3, we detect a possible adversarial sample when the two classifiers predict different classes.

\subsection{CW-$L_{\infty}$ attack}
\label{sec-resnet-vgg-cwinf}

\begin{figure}
  \centering
  \includegraphics[width=1.0\linewidth]{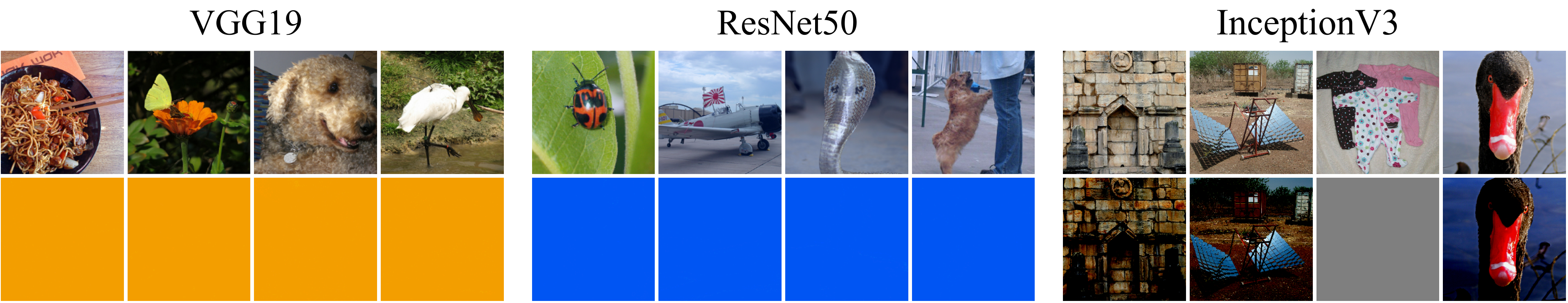}
  \caption{Here, we show CW-$L_{\infty}$ adversarial examples generated using VGG19, Resnet50 and InceptionV3 classifiers. These adversarial examples show that the adversarial examples depend on both input pre-processing and what attacks do. Because $L_{\infty}$ attack drives pixel values to a small $0.0-1.0$ range, this translates to blank VGG19 and Resnet50 adversarial images where the pre-processing image range is much bigger. For InceptionV3 images where pixels range from $-1.0$ to $1.0$, the CW-$L_{\infty}$ attack does not reduce all original images to blank images.}
  \label{fig:cwinf_vgg_resnet_incv3}
\end{figure}

From a first look at Table~\ref{tbl-imagenet-untargeted}, it would appear that these attacks are highly transferable, with close to $100$\% transfer rates in all classifiers regardless of where they were generated. However, taking a closer look at the adversarial images, we see that they are blank images that cause misclassification for the obvious reason that they are blank images. Figure~\ref{fig:cwinf_vgg_resnet_incv3} shows some adversarial examples generated with the CW-$L_{\infty}$ attack using all three classifiers. We can see that the ResNet50 and the VGG19 images are blank and rightfully misclassified by the classifiers. The attacks in this case transfer to other classifiers because they are blank.

Looking at how pixels change from CW-$L_{\infty}$ attack in VGG19 and Resnet50, many pixels change into $0.0$. As for the norms of channel minimums and maximums, the norms of channel minimums converge to $0.0$, whereas the norms of channel maximums converge to $1.0$. We think that the effect of the CW-$L_{\infty}$ attack depends on two things. On the one hand, the CW-$L_{\infty}$ attack seems to be clipping pixel values to the $0.0-1.0$ range. On the other hand, the range of image pixels in VGG19 and Resnet50 is much bigger than the range $0.0-1.0$ where the pixels are converged to by the attack. This leads CW-$L_{\infty}$ adversarial images for VGG19 and Resnet50 to be largely blank.

CW-$L_{\infty}$ are generated with IBM Art, as explained in . 
IBM art CW-$L_{\infty}$ attack does not take the clipping range as input but calculates it. This clearly affects the adversarial examples generated with IBM Art CW-$L_{\infty}$ attack.

\textbf{Detection of CW-$L_{\infty}$ adversarial examples} These adversarial images can easily be detected because many of the pixels become $0.0$.

\section{Examination of pixel values of InceptionV3 adversarial examples}
\label{sec-inc-exp}

In this section, we look closer at adversarial examples generated using the InceptionV3 classifier pre-trained on ImageNet.

\subsection{PGD and FGSM attacks}
\label{sec-incv3-pgd-fgsm}

We show PGD adversarial images generated with different $\epsilon$ values in Figure~\ref{fig:incv3_pgd}. Looking at transfer rates of PGD in Table~\ref{tbl-imagenet-untargeted} and Table~\ref{tbl-imagenet-targeted}, we can see that PGD attacks on InceptionV3 have 100\% transfer rates to other classifiers for $\epsilon=8$ and $\epsilon=16$. However, these high transfer rates have no bearing on the attack strength because we can see in Figure~\ref{fig:incv3_pgd} that the corresponding adversarial examples are simply noise.

Focusing on adversarial examples at pixel level, the PGD attack causes channel minimums and maximums to shift to the right towards the values $0.0$ or $1.0$ respectively. For non-adversarial images, channel minimums and maximums are close to $-1.0$ and $1.0$ respectively. The shift of channel minimums and maximums towards $0.0$ or $1.0$ is proportional to $\epsilon$ and converges to $0.0$ or $1.0$ starting from $\epsilon=1.0$. Table~\ref{tbl-pgd-shift} shows this shift in channel minimums and maximums. FGSM attacks behave the same as PGD attacks.

\begin{table*}[ht]
  \caption{Here, we show that in PGD adversarial examples in InceptionV3, averages of channel-wise minimum and maximum values shift to the right towards $0.0$ and $1.0$ respectively, converging there for $\epsilon>=1.0$.}
  \label{tbl-pgd-shift}
  \centering
  \begin{tabular}{lrrrrrr}
    \toprule
    \multicolumn{1}{c}{Images} & \multicolumn{2}{c}{Channel 0 avg.} & \multicolumn{2}{c}{Channel 1 avg.} & \multicolumn{2}{c}{Channel 2 avg.} \\
           \cmidrule(r){2-3} \cmidrule(r){4-5} \cmidrule(r){6-7}
            & min.     & max.    & min.    & max.    & min.    & max.    \\
    \midrule
    Original images               & -0.979 & 0.967 & -0.979 & 0.963 & -0.989 & 0.949 \\
    \midrule
    Adversarial (untarg PGD $\epsilon=8/127.5$) & -0.916 & 0.978 & -0.915 & 0.975 & -0.922 & 0.960 \\
    Adversarial (targ PGD $\epsilon=16/127.5$)  & -0.850 & 0.984 & -0.847 & 0.982 & -0.859 & 0.968 \\
    Adversarial (untarg PGD $\epsilon=1$)     &  0.000 & 0.999 &  0.000 & 1.000 &  0.000 & 0.999 \\
    Adversarial (targ PGD $\epsilon=2$)       &  0.000 & 1.000 &  0.000 & 1.000 &  0.000 & 1.000 \\
    \bottomrule
  \end{tabular}
\end{table*}

\textbf{Detection of CW-$L_{\infty}$ adversarial examples} Based on the shift of channel minimum and maximum values to the right towards $0.0$ or $1.0$ respectively, we can devise ways to detect PGD examples and original samples. For example, focusing on just one channel to detect PGD examples generated with an attack equal or stronger than a given $\epsilon$, we can use $\epsilon/2$ as a threshold on the average of channel minimum and channel maximum. We use a threshold on the average of the channel minimum and maximum because as these values shift in the same direction towards $0.0$ or $1.0$ respectively, their average also shifts to the right. We conduct such an experiment with channel 0, $\epsilon=8/127.5$, and a threshold of $\epsilon/2$. With this threshold, we get 88\% true negatives out of $1000$ random original images; find 90\% true positive adversarial images generated with PGD $\epsilon=8/127.5$, and find 97\% true positive adversarial images generated with PGD $\epsilon=16/127.5$. Then we conduct the same experiment with same channel 0 and $\epsilon=16/127.5$ to detect non-adversarial examples and adversarial examples generated with PGD attack equal or stronger than $\epsilon=16/127.5$. We find 91\% true negatives out of $1000$ random original images, and find 95\% true positive adversarial images generated with PGD $\epsilon=16/127.5$.

\subsection{CW-$L_2$ attack}

From Table\ref{tbl-imagenet-untargeted}, we see that untargeted CW-$L_2$ attacks do not transfer to the ResNet50 and VGG19 clssifiers.

\textbf{Detection of CW-$L_2$ adversarial examples} We can use another classifier of ResNet50 or VGG19 architecture to detect adversarial examples because the attacks do not transfer to these other architectures.

\subsection{CW-$L_{\infty}$ attack}
\label{sec-incv3-cwinf}

Looking at how pixels change from CW-$L_{\infty}$ attack in InceptionV3, positive value pixels stay the same, negative pixels change into $0.0$. As for the norms of channel minimums and maximums, the norms of channel minimums converge to $0.0$, whereas the norms of channel maximums converge to $1.0$ or values slightly below $1.0$. Occasionally, all channel minimums and maximums collapse to $0.0$, which leads to blank adversarial images such as the third CW-$L_{\infty}$ attack image for InceptionV3 in Figure~\ref{fig:cwinf_vgg_resnet_incv3}.

\textbf{Detection of CW-$L_{\infty}$ adversarial examples} These adversarial images can be detected because many of the pixels become $0$. We run an experiment to count the number of $0.0$ pixels in CW-$L_{\infty}$ adversarial examples. We start with $1000$ random ImageNet images and generate CW-$L_{\infty}$ adversarial examples from them. Then we look at the samples the were classified correctly as original images, but misclassified as adversarial images. We calculate the number of $0$ pixels in original examples and their corresponding adversarial images. The results in Table~\ref{tbl-cwinf-zeros} show that the average number of pixels with value exactly $0.0$ increases from fewer than 10 per image to up to hundreds of thousands. The measurements are done on images on their InceptionV3 pre-processing format.

\begin{table*}[ht]
  \caption{Here, we show that the number of $0.0$ pixels in adversarial images increases substantially in CW-$L_{\infty}$ adversarial examples, in comparison to their corresponding original images.}
  \label{tbl-cwinf-zeros}
  \centering
  \begin{tabular}{lrrrrrrrr}
    \toprule
    \multicolumn{1}{c}{Classifier} & \multicolumn{4}{c}{Original images} & \multicolumn{4}{c}{Adversarial images} \\
           \cmidrule(r){2-5} \cmidrule(r){6-9} 
               & Min. \# & Avg. \#  & Max. \#  & St.dev. \# & Min. \#  & Avg. \#  & Max. \#  & St.dev. \# \\
               & of zero & of zero  & of zero  & of zero    & of zero  & of zero  & of zero  & of zero    \\
    Classifier & pixels  & pixels   & pixels   & pixels     & pixels   & pixels   & pixels   & pixels     \\
    \midrule
    InceptionV3 & 0 & 0.1 & 5 & 0.5 &  14778 & 178270.6 & 268203 & 63102.4 \\
    ResNet50    & 0 & 0 0 & 1 & 0.1 & 145576 & 149604.3 & 150486 &   621.4 \\
    VGG19       & 0 & 0.0 & 1 & 0.1 & 146171 & 149654.5 & 150512 &   512.3 \\
    \bottomrule
  \end{tabular}
\end{table*}

\section{Broader impact}
\label{sec-discussion}

The attacks we consider are among the strongest current attacks. A possible limitation could be that other attacks might behave differently. We believe our research does not have potential negative societal impact. On the contrary, we think its broader impact would be positive because it would contribute to the safety and security of deep learning solutions by enabling adversarial defenses to counter adversarial attacks.

\section{Conclusions}

We have shown instances of strong attacks with low transferability among undefended classifiers. This is despite the fact that adversarial attacks are claimed to be highly transferable. We also find that high attack transfer rates can be misleading as a metric of attack strength. The reason is that attacks that generate out-of-domain adversarial images also have high transfer rate simply because such images misclassify across classifiers. From closer examination of image pixels, we conclude that image pre-processing in classifiers affects attack impact. From insights gained from examination of pixel changes from attacks, we are able to devise simple thresholds to tell apart adversarial samples and original samples.

\bibliography{egbib}

\begin{thebibliography}{33}
\providecommand{\natexlab}[1]{#1}
\providecommand{\url}[1]{\texttt{#1}}
\expandafter\ifx\csname urlstyle\endcsname\relax
  \providecommand{\doi}[1]{doi: #1}\else
  \providecommand{\doi}{doi: \begingroup \urlstyle{rm}\Url}\fi

\bibitem[Athalye et~al.(2018)Athalye, Carlini, and
  Wagner]{athalye2018obfuscated}
Anish Athalye, Nicholas Carlini, and David Wagner.
\newblock Obfuscated gradients give a false sense of security: Circumventing
  defenses to adversarial examples.
\newblock \emph{arXiv preprint arXiv:1802.00420}, 2018.

\bibitem[Carlini \& Wagner(2016)Carlini and Wagner]{carlini2016defensive}
Nicholas Carlini and David Wagner.
\newblock Defensive distillation is not robust to adversarial examples.
\newblock \emph{arXiv preprint arXiv:1607.04311}, 2016.

\bibitem[Carlini \& Wagner(2017{\natexlab{a}})Carlini and
  Wagner]{carlini2017adversarial}
Nicholas Carlini and David Wagner.
\newblock Adversarial examples are not easily detected: Bypassing ten detection
  methods.
\newblock In \emph{Proceedings of the 10th ACM Workshop on Artificial
  Intelligence and Security}, pp.\  3--14. ACM, 2017{\natexlab{a}}.

\bibitem[Carlini \& Wagner(2017{\natexlab{b}})Carlini and
  Wagner]{carlini2017magnet}
Nicholas Carlini and David Wagner.
\newblock Magnet and" efficient defenses against adversarial attacks" are not
  robust to adversarial examples.
\newblock \emph{arXiv preprint arXiv:1711.08478}, 2017{\natexlab{b}}.

\bibitem[Carlini \& Wagner(2017{\natexlab{c}})Carlini and
  Wagner]{carlini2017towards}
Nicholas Carlini and David Wagner.
\newblock Towards evaluating the robustness of neural networks.
\newblock In \emph{2017 IEEE Symposium on Security and Privacy (SP)}, pp.\
  39--57. IEEE, 2017{\natexlab{c}}.

\bibitem[Carlini et~al.(2019)Carlini, Athalye, Papernot, Brendel, Rauber,
  Tsipras, Goodfellow, Madry, and Kurakin]{carlini2019evaluating}
Nicholas Carlini, Anish Athalye, Nicolas Papernot, Wieland Brendel, Jonas
  Rauber, Dimitris Tsipras, Ian Goodfellow, Aleksander Madry, and Alexey
  Kurakin.
\newblock On evaluating adversarial robustness.
\newblock \emph{arXiv preprint arXiv:1902.06705}, 2019.

\bibitem[Chen et~al.(2017)Chen, Zhang, Sharma, Yi, and Hsieh]{chen2017zoo}
Pin-Yu Chen, Huan Zhang, Yash Sharma, Jinfeng Yi, and Cho-Jui Hsieh.
\newblock Zoo: Zeroth order optimization based black-box attacks to deep neural
  networks without training substitute models.
\newblock In \emph{Proceedings of the 10th ACM Workshop on Artificial
  Intelligence and Security}, pp.\  15--26. ACM, 2017.

\bibitem[Chollet et~al.(2015)]{chollet2015keras}
Fran\c{c}ois Chollet et~al.
\newblock Keras.
\newblock \url{https://keras.io}, 2015.

\bibitem[Chollet(2017)]{chollet2017xception}
Fran{\c{c}}ois Chollet.
\newblock Xception: Deep learning with depthwise separable convolutions.
\newblock In \emph{Proceedings of the IEEE conference on computer vision and
  pattern recognition}, pp.\  1251--1258, 2017.

\bibitem[Goodfellow et~al.(2014)Goodfellow, Shlens, and
  Szegedy]{goodfellow6572explaining}
Ian~J Goodfellow, Jonathon Shlens, and Christian Szegedy.
\newblock Explaining and harnessing adversarial examples.
\newblock \emph{arXiv preprint arXiv:1412.6572}, 2014.

\bibitem[Guo et~al.(2020)Guo, Li, and Chen]{NEURIPS2020_00e26af6}
Yiwen Guo, Qizhang Li, and Hao Chen.
\newblock Backpropagating linearly improves transferability of adversarial
  examples.
\newblock In H.~Larochelle, M.~Ranzato, R.~Hadsell, M.~F. Balcan, and H.~Lin
  (eds.), \emph{Advances in Neural Information Processing Systems}, volume~33,
  pp.\  85--95. Curran Associates, Inc., 2020.
\newblock URL
  \url{https://proceedings.neurips.cc/paper/2020/file/00e26af6ac3b1c1c49d7c3d79c60d000-Paper.pdf}.

\bibitem[He et~al.(2016)He, Zhang, Ren, and Sun]{he2016deep}
Kaiming He, Xiangyu Zhang, Shaoqing Ren, and Jian Sun.
\newblock Deep residual learning for image recognition.
\newblock In \emph{Proceedings of the IEEE conference on computer vision and
  pattern recognition}, pp.\  770--778, 2016.

\bibitem[Huang et~al.(2019)Huang, Katsman, He, Gu, Belongie, and
  Lim]{huang2019enhancing}
Qian Huang, Isay Katsman, Horace He, Zeqi Gu, Serge Belongie, and Ser-Nam Lim.
\newblock Enhancing adversarial example transferability with an intermediate
  level attack.
\newblock In \emph{Proceedings of the IEEE/CVF International Conference on
  Computer Vision}, pp.\  4733--4742, 2019.

\bibitem[Ilyas et~al.(2019)Ilyas, Santurkar, Tsipras, Engstrom, Tran, and
  Madry]{ilyas2019adversarial}
Andrew Ilyas, Shibani Santurkar, Dimitris Tsipras, Logan Engstrom, Brandon
  Tran, and Aleksander Madry.
\newblock Adversarial examples are not bugs, they are features.
\newblock \emph{arXiv preprint arXiv:1905.02175}, 2019.

\bibitem[Inkawhich et~al.(2019)Inkawhich, Wen, Li, and
  Chen]{inkawhich2019feature}
Nathan Inkawhich, Wei Wen, Hai~Helen Li, and Yiran Chen.
\newblock Feature space perturbations yield more transferable adversarial
  examples.
\newblock In \emph{Proceedings of the IEEE/CVF Conference on Computer Vision
  and Pattern Recognition}, pp.\  7066--7074, 2019.

\bibitem[Inkawhich et~al.(2020{\natexlab{a}})Inkawhich, Liang, Wang, Inkawhich,
  Carin, and Chen]{NEURIPS2020_eefc7bfe}
Nathan Inkawhich, Kevin Liang, Binghui Wang, Matthew Inkawhich, Lawrence Carin,
  and Yiran Chen.
\newblock Perturbing across the feature hierarchy to improve standard and
  strict blackbox attack transferability.
\newblock In H.~Larochelle, M.~Ranzato, R.~Hadsell, M.~F. Balcan, and H.~Lin
  (eds.), \emph{Advances in Neural Information Processing Systems}, volume~33,
  pp.\  20791--20801. Curran Associates, Inc., 2020{\natexlab{a}}.
\newblock URL
  \url{https://proceedings.neurips.cc/paper/2020/file/eefc7bfe8fd6e2c8c01aa6ca7b1aab1a-Paper.pdf}.

\bibitem[Inkawhich et~al.(2020{\natexlab{b}})Inkawhich, Liang, Carin, and
  Chen]{inkawhich2020transferable}
Nathan Inkawhich, Kevin~J Liang, Lawrence Carin, and Yiran Chen.
\newblock Transferable perturbations of deep feature distributions.
\newblock In \emph{ICLR}, 2020{\natexlab{b}}.

\bibitem[Kurakin et~al.(2016)Kurakin, Goodfellow, and
  Bengio]{kurakin2016adversarial}
Alexey Kurakin, Ian Goodfellow, and Samy Bengio.
\newblock Adversarial machine learning at scale.
\newblock \emph{arXiv preprint arXiv:1611.01236}, 2016.

\bibitem[Kurakin et~al.(2018)Kurakin, Goodfellow, Bengio, Dong, Liao, Liang,
  Pang, Zhu, Hu, Xie, et~al.]{kurakin2018adversarial}
Alexey Kurakin, Ian Goodfellow, Samy Bengio, Yinpeng Dong, Fangzhou Liao, Ming
  Liang, Tianyu Pang, Jun Zhu, Xiaolin Hu, Cihang Xie, et~al.
\newblock Adversarial attacks and defences competition.
\newblock In \emph{The NIPS'17 Competition: Building Intelligent Systems}, pp.\
   195--231. Springer, 2018.

\bibitem[Madry et~al.(2018)Madry, Makelov, Schmidt, Tsipras, and
  Vladu]{madry2017towards}
Aleksander Madry, Aleksandar Makelov, Ludwig Schmidt, Dimitris Tsipras, and
  Adrian Vladu.
\newblock Towards deep learning models resistant to adversarial attacks.
\newblock In \emph{ICLR}, 2018.

\bibitem[Moosavi-Dezfooli et~al.(2016)Moosavi-Dezfooli, Fawzi, and
  Frossard]{moosavi2016deepfool}
Seyed-Mohsen Moosavi-Dezfooli, Alhussein Fawzi, and Pascal Frossard.
\newblock Deepfool: a simple and accurate method to fool deep neural networks.
\newblock In \emph{Proceedings of the IEEE conference on computer vision and
  pattern recognition}, pp.\  2574--2582, 2016.

\bibitem[Nicolae et~al.(2018)Nicolae, Sinn, Tran, Buesser, Rawat, Wistuba,
  Zantedeschi, Baracaldo, Chen, Ludwig, Molloy, and Edwards]{art2018}
Maria-Irina Nicolae, Mathieu Sinn, Minh~Ngoc Tran, Beat Buesser, Ambrish Rawat,
  Martin Wistuba, Valentina Zantedeschi, Nathalie Baracaldo, Bryant Chen, Heiko
  Ludwig, Ian Molloy, and Ben Edwards.
\newblock Adversarial robustness toolbox v1.2.0.
\newblock \emph{CoRR}, 1807.01069, 2018.
\newblock URL \url{https://arxiv.org/pdf/1807.01069}.

\bibitem[Papernot et~al.(2016)Papernot, McDaniel, Wu, Jha, and
  Swami]{papernot2016distillation}
Nicolas Papernot, Patrick McDaniel, Xi~Wu, Somesh Jha, and Ananthram Swami.
\newblock Distillation as a defense to adversarial perturbations against deep
  neural networks.
\newblock In \emph{2016 IEEE Symposium on Security and Privacy (SP)}, pp.\
  582--597. IEEE, 2016.

\bibitem[Russakovsky et~al.(2015)Russakovsky, Deng, Su, Krause, Satheesh, Ma,
  Huang, Karpathy, Khosla, Bernstein, Berg, and Fei-Fei]{ILSVRC15}
Olga Russakovsky, Jia Deng, Hao Su, Jonathan Krause, Sanjeev Satheesh, Sean Ma,
  Zhiheng Huang, Andrej Karpathy, Aditya Khosla, Michael Bernstein,
  Alexander~C. Berg, and Li~Fei-Fei.
\newblock {ImageNet Large Scale Visual Recognition Challenge}.
\newblock \emph{International Journal of Computer Vision (IJCV)}, 115\penalty0
  (3):\penalty0 211--252, 2015.
\newblock \doi{10.1007/s11263-015-0816-y}.

\bibitem[Sabour et~al.(2016)Sabour, Cao, Faghri, and
  Fleet]{sabour2015adversarial}
Sara Sabour, Yanshuai Cao, Fartash Faghri, and David~J Fleet.
\newblock Adversarial manipulation of deep representations.
\newblock In \emph{International Conference on Learning Representations}, 2016.

\bibitem[Simonyan \& Zisserman(2015)Simonyan and Zisserman]{simonyan2015very}
Karen Simonyan and Andrew Zisserman.
\newblock Very deep convolutional networks for large-scale image recognition.
\newblock In Yoshua Bengio and Yann LeCun (eds.), \emph{3rd International
  Conference on Learning Representations, {ICLR} 2015, San Diego, CA, USA, May
  7-9, 2015, Conference Track Proceedings}, 2015.
\newblock URL \url{http://arxiv.org/abs/1409.1556}.

\bibitem[Szegedy et~al.(2013)Szegedy, Zaremba, Sutskever, Bruna, Erhan,
  Goodfellow, and Fergus]{szegedy2013intriguing}
Christian Szegedy, Wojciech Zaremba, Ilya Sutskever, Joan Bruna, Dumitru Erhan,
  Ian~J. Goodfellow, and Rob Fergus.
\newblock Intriguing properties of neural networks.
\newblock In \emph{International Conference on Learning Representations}, 2013.

\bibitem[Szegedy et~al.(2016)Szegedy, Vanhoucke, Ioffe, Shlens, and
  Wojna]{szegedy2016rethinking}
Christian Szegedy, Vincent Vanhoucke, Sergey Ioffe, Jon Shlens, and Zbigniew
  Wojna.
\newblock Rethinking the inception architecture for computer vision.
\newblock In \emph{Proceedings of the IEEE conference on computer vision and
  pattern recognition}, pp.\  2818--2826, 2016.

\bibitem[Szegedy et~al.(2017)Szegedy, Ioffe, Vanhoucke, and
  Alemi]{szegedy2017inception}
Christian Szegedy, Sergey Ioffe, Vincent Vanhoucke, and Alexander Alemi.
\newblock Inception-v4, inception-resnet and the impact of residual connections
  on learning.
\newblock In \emph{Proceedings of the AAAI Conference on Artificial
  Intelligence}, volume 31 Issue 1, 2017.

\bibitem[Tan \& Le(2019)Tan and Le]{tan2019efficientnet}
Mingxing Tan and Quoc Le.
\newblock Efficientnet: Rethinking model scaling for convolutional neural
  networks.
\newblock In \emph{International Conference on Machine Learning}, pp.\
  6105--6114. PMLR, 2019.

\bibitem[Tram{\`e}r et~al.(2017)Tram{\`e}r, Papernot, Goodfellow, Boneh, and
  McDaniel]{tramer2017space}
Florian Tram{\`e}r, Nicolas Papernot, Ian Goodfellow, Dan Boneh, and Patrick
  McDaniel.
\newblock The space of transferable adversarial examples.
\newblock \emph{arXiv preprint arXiv:1704.03453}, 2017.

\bibitem[Tramer et~al.(2020)Tramer, Carlini, Brendel, and
  Madry]{tramer2020adaptive}
Florian Tramer, Nicholas Carlini, Wieland Brendel, and Aleksander Madry.
\newblock On adaptive attacks to adversarial example defenses.
\newblock \emph{arXiv preprint arXiv:2002.08347}, 2020.

\bibitem[Zhou et~al.(2018)Zhou, Hou, Chen, Tang, Huang, Gan, and
  Yang]{zhou2018transferable}
Wen Zhou, Xin Hou, Yongjun Chen, Mengyun Tang, Xiangqi Huang, Xiang Gan, and
  Yong Yang.
\newblock Transferable adversarial perturbations.
\newblock In \emph{Proceedings of the European Conference on Computer Vision
  (ECCV)}, pp.\  452--467, 2018.

\end{thebibliography}
\bibliographystyle{iclr2021_conference}

\newpage

\appendix

\section{Appendix}

\begin{table*}[ht]
  \caption{Similarly to Table~\ref{tbl-imagenet-untargeted}, here we show transfer rates and average perturbation for targeted attacks on ImageNet classifiers. (-) indicates that transfer rate could not be measured based on the definition of transfer rate because there was no original sample that classified correctly, the adversarial example of which became adversarial. Targeted attacks seem to be not as strong as untargeted attacks. The three observations we made for untargeted attacks hold for targeted attacks also. (1) The attacks with highest transfer rate are the ones that cause close to maximum adversarial perturbation that classifier pixel range allows ($2.0$ for InceptionV3 and $255.0$ for ResNet50 and VGG19). However, the high transfer rates fail to capture the fact that many of the adversarial examples that transfer to other classifiers are out-of-domain, just-noise images that naturally misclassify. The high perturbation values are an indication that the generated adversarial examples do not look like domain images. (2) The transfer rates of strong attacks such as CW and PGD can be as low as $17$\%. (3) For VGG19, adversarial perturbation seems to saturate to values well below the maximum that the $255.0$ pixel range should allow. The value where adversarial perturbation seems to saturate to is an indication that adversarial perturbation may depend on input pre-processing.}
  \label{tbl-imagenet-targeted}
  \centering
  \begin{tabular}{llrrrr}
    \toprule
          \multicolumn{2}{c}{Attack}  & \multicolumn{1}{c}{Perturbation} & \multicolumn{3}{c}{Transfer rate} \\
                                        \cmidrule(r){1-2}   \cmidrule(r){3-3} \cmidrule(r){4-6}
    Attacked     &                   & Avg. $L_{\infty}$   & & & \\
    model       & Attack            & distances & InceptionV3 & VGG19  & ResNet50 \\

    \midrule

    InceptionV3 & PGD $\epsilon=16$       & 1.99 & 100\% & 100\% & 100\%  \\ 
    InceptionV3 & PGD $\epsilon=16/127.5$ & 0.13 & 100\% &  18\%  &  17\% \\ 
    InceptionV3 & CW-$L_2,\kappa=0$    &    - &  -    & -             & -      \\ 
    InceptionV3 & CW-$L_2,\kappa=100$  &    - & -     & -             & -      \\ 
    InceptionV3 & CW-$L_{\infty}$      & 0.96 & 100\% & 100\%         & 100\%  \\  

    \midrule
    
    VGG19 & PGD $\epsilon=16$            &   16.00 & 30\% & 100\% & 41\% \\ 
    VGG19 & CW-$L_2, \kappa=0$         &  - & - & - & - \\ 
    VGG19 & CW-$L_2, \kappa=100$       &  - & - & - & - \\ 
    VGG19 & CW-$L_{\infty}$            & 150.79 & 100\% & 100\% & 100\% \\ 

    \midrule
    
    ResNet50 & PGD $\epsilon=16$      &    16.00 & 36\% & 63\% & 100\% \\ 
    ResNet50 & CW-$L_2, \kappa=0$   &  - & - & - & -   \\ 
    ResNet50 & CW-$L_2, \kappa=100$ &  - & - & - & -   \\ 
    ResNet50 & CW-$L_{\infty}$      &  - &      - &      - &       - \\ 
    \bottomrule
  \end{tabular}
\end{table*}

\begin{table*}[ht]
  \caption{Here, we show channel-wise, pixel minimum, average, maximum, standard deviation for 1000 random, original ImageNet images. Based on average and standard deviation values, we can see that the channel minimum and maximum values are concentrated near extremum values.}
  \label{tbl-pgd-min-max}
  \centering
  \begin{tabular}{lrrrrrr}
    \toprule
    \multicolumn{1}{c}{Metrics} & \multicolumn{2}{c}{Channel 0} & \multicolumn{2}{c}{Channel 1} & \multicolumn{2}{c}{Channel 2} \\
           \cmidrule(r){2-3} \cmidrule(r){4-5} \cmidrule(r){6-7}
    Metrics & Channel     & Channel   & Channel     & Channel   & Channel     & Channel   \\
    Metrics & minimum     & maximum   & minimum     & maximum   & minimum     & maximum   \\
    \midrule
    limit   &     -103.939 & 151.061 &  116.779 & 138.221 & -123.680 & 131.320 \\
    \midrule
    minimum &     -103.939 &   9.487 & -116.779 &  -0.869 & -123.680 &   9.242 \\
    average &     -101.169 & 141.940 & -112.059 & 131.574 & -119.012 & 125.397 \\
    maximum &      -35.985 & 151.061 &  -14.248 & 138.221 &   10.679 & 131.320 \\
    standard dev. &  7.071 &  18.517 &    9.091 &  14.501 &   10.617 &  13.575 \\
    \bottomrule
  \end{tabular}
\end{table*}

\begin{figure}[h]
  \centering
  \includegraphics[width=1.0\linewidth]{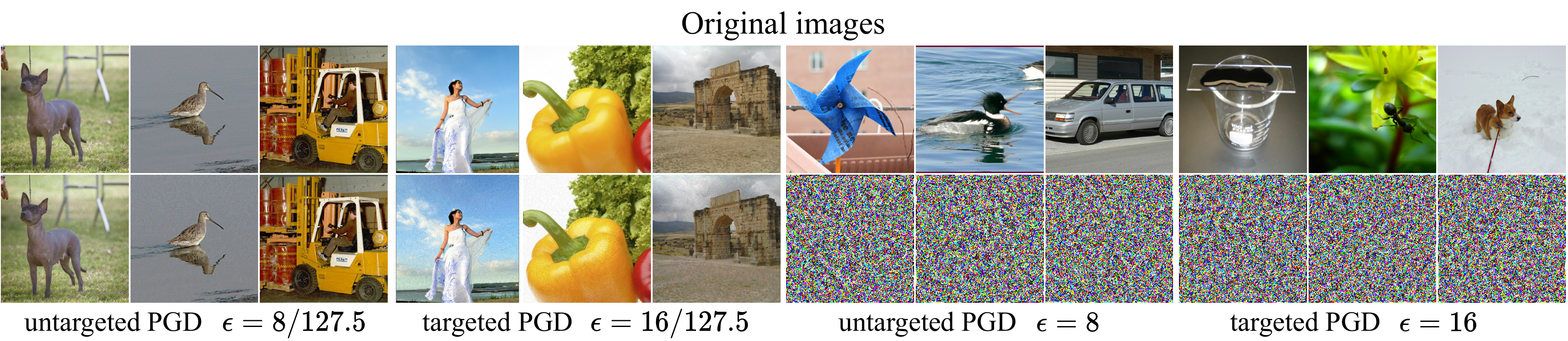}
  \caption{Here, we show PGD adversarial images generated with different $\epsilon$ values with InceptionV3 classifier. The images on the right have 100\% transferability rate to other classifiers based on Table~\ref{tbl-imagenet-untargeted} and Table~\ref{tbl-imagenet-targeted}. However, we can see that for the images on the right, the adversarial images are just noise. Therefore, transfer rate of attacks does not in this case indicate a strong attack.}
  \label{fig:incv3_pgd}
\end{figure}

\end{document}